\documentclass[10pt,twocolumn,letterpaper]{article}

\usepackage[pagenumbers]{cvpr} 
\usepackage{xcolor}         
\usepackage{graphicx}
\usepackage{bbding}
\usepackage{amsmath}
\usepackage{multirow}
\usepackage{tabularx}
\usepackage{float}  
\definecolor{cvprblue}{rgb}{0.21,0.49,0.74}
\usepackage[pagebackref,breaklinks,colorlinks,allcolors=cvprblue]{hyperref}

\title{MiM-DiT: MoE in MoE with Diffusion Transformers \\ for All-in-One Image Restoration }

\author{%
    Lingshun Kong$^1$, Jiawei Zhang, Zhengpeng Duan$^2$, Xiaohe Wu$^3$\\
    Yueqi Yang, Xiaotao Wang, Dongqing Zou, Lei Lei, Jinshan Pan$^1$\\
    $^1$Nanjing University of Science and Technology \\
    $^2$Nankai University $^3$Harbin Institute of Technology\\
}

\begin{document}
\maketitle
\begin{abstract}

All-in-one image restoration is challenging because different degradation types, such as haze, blur, noise, and low-light, impose diverse requirements on restoration strategies, making it difficult for a single model to handle them effectively.
In this paper, we propose a unified image restoration framework that integrates a dual-level Mixture-of-Experts (MoE) architecture with a pretrained diffusion model. 
The framework operates at two levels: the Inter-MoE layer adaptively combines expert groups to handle major degradation types, while the Intra-MoE layer further selects specialized sub-experts to address fine-grained variations within each type.
This design enables the model to achieve coarse-grained adaptation across diverse degradation categories while performing fine-grained modulation for specific intra-class variations, ensuring both high specialization in handling complex, real-world corruptions.
Extensive experiments demonstrate that the proposed method performs favorably against the state-of-the-art
approaches on multiple image restoration tasks.
\end{abstract}

\section{Introduction}

%
All-in-one image restoration aims to reconstruct high-quality images from variably degraded inputs; however, it faces a fundamental challenge: the model must handle multiple degradation types that impose divergent structural requirements. 
This is because reversing some degradations depends on leveraging global image statistics, while addressing others necessitates the reconstruction of sharp local structures, thus imposing conflicting modeling priorities on a unified architecture.
These inherent conflicts demand a restoration system capable of dynamically adapting its processing strategy to the specific characteristics of each input degradation without compromising structural fidelity or perceptual quality.
\begin{figure}[!t]
    \centering
 \includegraphics[width=0.48\textwidth]{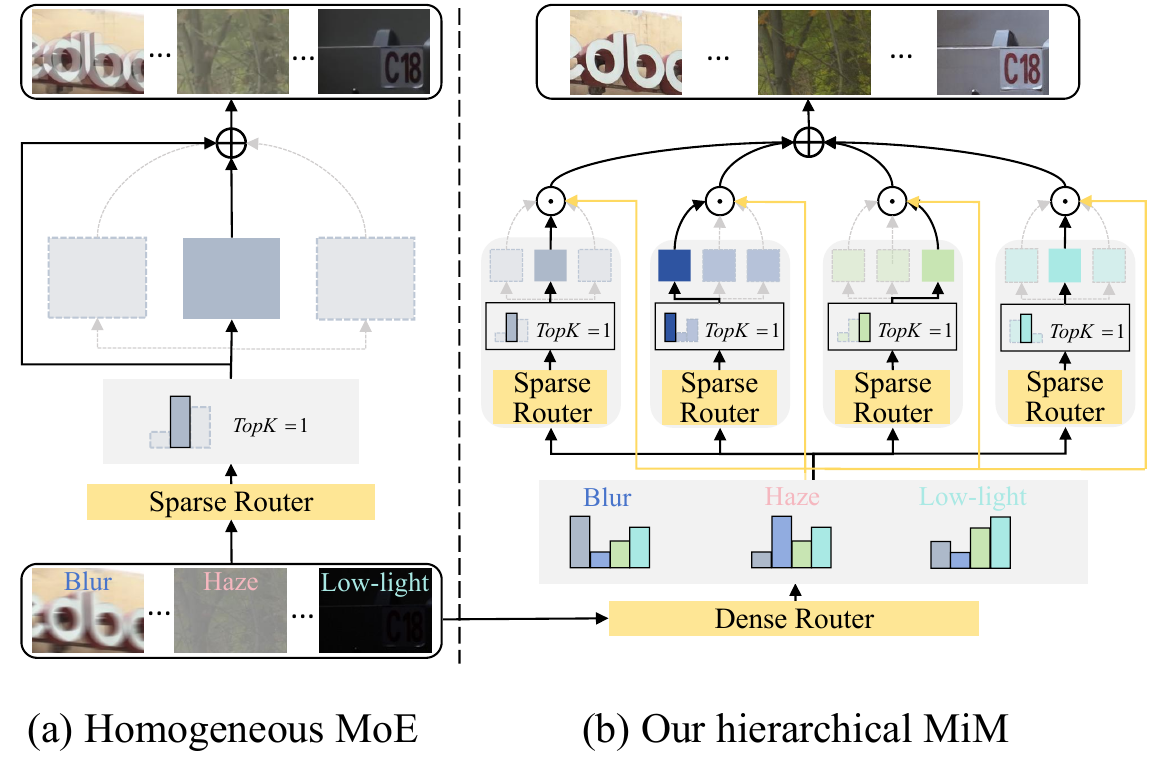} 
 \vspace{-6mm}
\caption{Homogeneous MoE \emph{vs.} the proposed MiM. 
The symbols in the figure are consistent with those defined in our method section and share the same meanings.
Our hierarchical MiM adopts a dual-level routing mechanism, which ensures dynamic and specialized processing by routing inputs to appropriate architectural priors and fine-grained specialists.
Compared with homogeneous MoE, our method enables better restoration for blur, haze, and low-light conditions through adaptive structural selection.
}
 \vspace{-6mm}
 \label{fig: mov}
\end{figure}

To solve this problem, existing methods either use a general-purpose network architecture that is trained on different tasks separately~\cite{MPRNet,Restormer,NAFNet,dualCNN,physicgan,dualCNNconfernce} or employ unified models with shared parameters to handle multiple degradations~\cite{promptir,airnet}. 
However, the former requires task-specific training, while the latter produces generic results due to compromised parameter sharing. 
Although recent methods employ MoE to activate specialized experts dynamically~\cite{unirestorer,moceir}, they still rely on deterministic regression losses, leading to over-smoothed outputs and limited texture generation.

Diffusion models have demonstrated powerful generative capabilities in specialized tasks like super-resolution~\cite{diffbir,pasd,resshift,dit4sr}. 
Several works have extended them to all-in-one restoration~\cite{autodir, DiffUIR}. 
However, they typically apply a uniform diffusion process to all degradations, ignoring their distinct nature. This often leads to outputs with structural distortions or missing textual details.
While MoE-based methods produce over-smoothed results due to deterministic losses and diffusion models employ uniform sampling that ignores degradation-specific characteristics, combining their strengths appears promising.
However, simply integrating diffusion models with MoE frameworks remains suboptimal, as it retains homogeneous expert designs that overlook distinct inductive biases across tasks (Figure~\ref{fig: mov}(a)). For instance, a single attention mechanism struggles to concurrently model long-range spatial dependencies and global cross-channel interactions.
To address the aforementioned challenges, we propose a hierarchical MoE in MoE (MiM) architecture that seamlessly integrates the MoE paradigm with a pre-trained diffusion model.
Our approach tackles the limitation of homogeneous experts through a dual-level routing mechanism (Inter-MoE and Intra-MoE), which ensures dynamic and specialized processing by routing inputs to appropriate architectural priors and fine-grained specialists.
At the Inter-MoE level, we construct four expert groups based on distinct attention mechanisms: spatial self-attention~\cite{Transformer}, channel self-attention~\cite{Restormer}, Swin attention~\cite{Swin}, and SE attention~\cite{senet}. 
This set is chosen to capture a comprehensive suite of inductive biases spanning global and local spatial dependencies, as well as inter-channel relationships, that are essential for handling diverse restoration tasks.. 
All groups process the degraded image concurrently in the latent space. A dense router then adaptively integrates their outputs according to the input's degradation characteristics, enabling dynamic and context-aware feature fusion.
Unlike homogeneous models that apply identical processing to all degradations and often yield compromised performance (Figure~\ref{fig: mov}(a)), our method dynamically selects the most suitable architectural priors for each specific degradation, enabling better and more specialized restoration across diverse conditions (Figure~\ref{fig: mov}(b)).

Within each expert group, the Intra-MoE layer further deploys multiple sub-experts that share the same attention mechanism as their parent group but specialize in different intra-class degradation variations (e.g., light vs. heavy haze). 
A sparse router dynamically selects and activates the most relevant sub-experts for each input, ensuring both specialization and efficiency. 
Together, this dual-level hierarchy enables coarse-grained adaptation across structural model types and fine-grained modulation within each type. 
By embedding this hierarchical MoE-in-MoE design into a pre-trained diffusion backbone, our method achieves robust restoration across diverse real-world scenarios.

The main contributions are summarized as follows:
\begin{itemize}
    \item We propose an all-in-one image restoration framework that integrates the Mixture-of-Experts (MoE) paradigm with a pretrained diffusion model, enabling unified modeling of diverse degradation types while preserving the favorable generative capacity of the pre-trained backbone.
    \item We introduce a hierarchical MoE-in-MoE architecture comprising two complementary levels: an Inter-MoE that routes between experts of distinct architectures, and an Intra-MoE that routes to specialized sub-experts within each architecture. This hierarchical setup enables dynamic and granular adaptation to both inter-class and intra-class degradation patterns.
    \item We conduct extensive experiments showing that our method achieves state-of-the-art performance in various image restoration benchmarks.
\end{itemize}

\section{Related Work}
%
%
\begin{figure*}[!t]
    \centering
 \includegraphics[width=0.98\textwidth]{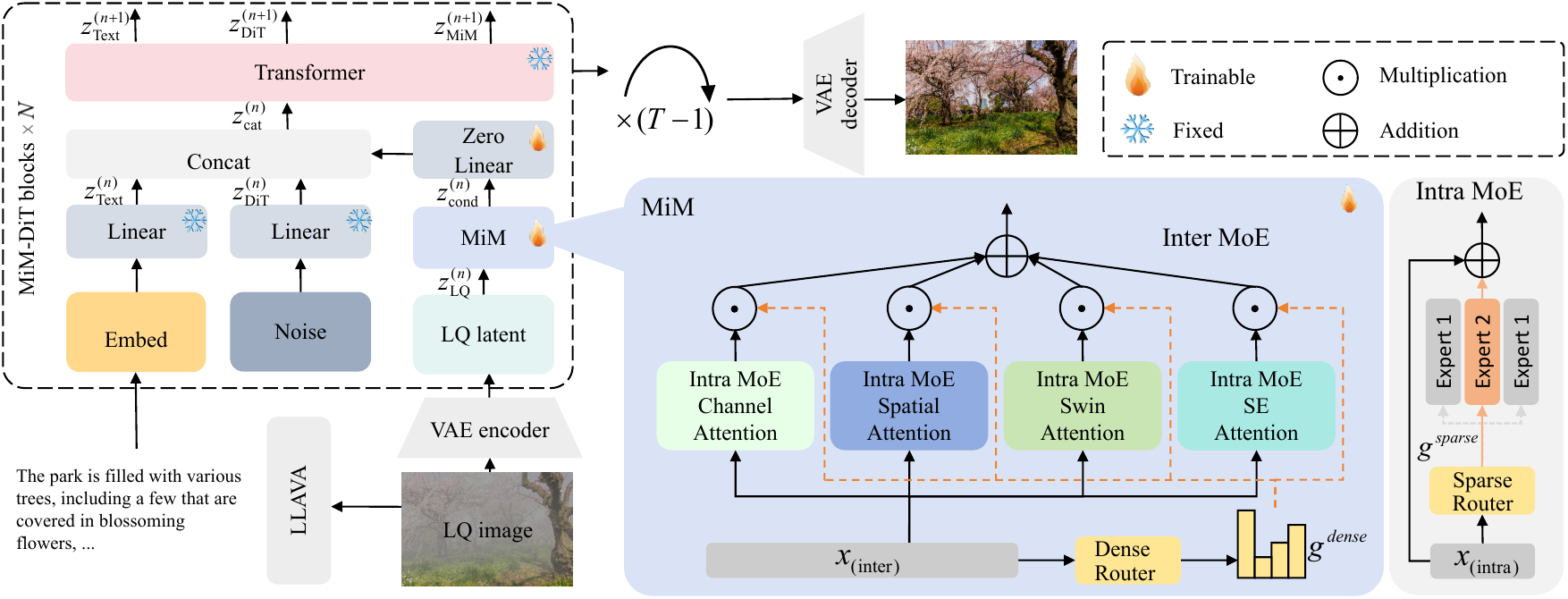} 
 \vspace{-3mm}
\caption{Overview of the proposed Hierarchical MoE in MoE (MiM) integrated into the DiT backbone. The framework processes low-quality (LQ) images through a series of MiM-DiT blocks. Given the LQ input, the MiM module extracts degradation-specific features and processes them through a hierarchical MoE architecture composed of two levels: Inter-MoE and Intra-MoE. At the Inter-MoE level, four expert groups based on distinct attention mechanisms—spatial self-attention channel, self-attention, Swin attention, and SE attention—are combined via a dense router that computes adaptive weights over all groups. This dense fusion enables the model to leverage complementary inductive biases. Within each expert group, Intra-MoE captures fine-grained variations within each degradation category via sparse routing. These processed features are then injected as conditional input into the DiT backbone through a Zero-Linear pathway, dynamically guiding the diffusion process to generate restoration results.}
 \vspace{-5mm}
 \label{fig: Network}
\end{figure*}

{\flushleft\textbf{All-in-one image restoration}.}
Recent advances in all-in-one image restoration aim to handle multiple degradation types within a unified framework~\cite{aio1}. 
Compared to task-specific methods~\cite{MPRNet, Restormer, cat}, this paradigm faces greater challenges due to the fundamental differences and sometimes mutually exclusive nature of various degradation processes. 
AirNet~\cite{airnet} introduced contrastive learning to extract discriminative degradation representations, while PromptIR~\cite{promptir} and ProRes~\cite{Prores} proposed using learnable prompts extracted from inputs to guide network behavior.
Recent research has explored integrating pretrained large-scale vision models~\cite{aio_lvm1} and multimodal large language models~\cite{DA-CLIP,instructir,uniprocessor} into backbone networks to facilitate high-fidelity universal image restoration. 
FoundIR~\cite{foundir} introduces an HD all-in-one dataset spanning diverse degradations for unified restoration training.
However, their reliance on deterministic regression and pixel-level losses limits their ability to model the ill-posed nature of restoration, often leading to over-smoothed results.

{\flushleft \textbf{Diffusion-based image restoration.}}
%
%
A number of methods~\cite{resshift,cdm,hidiff} have attempted to address image restoration by directly training diffusion models.
Building on this line of work, several recent approaches have leveraged large-scale pre-trained generative models such as Stable Diffusion (SD)~\cite{sd} to enhance restoration performance.
%
DiffBIR~\cite{diffbir} adopts a two-stage pipeline that first recovers the degraded image structurally and then employs SD to refine visual details.
Similarly, PASD~\cite{pasd} introduces a Degradation Removal module to produce clean conditional inputs, which are then fed into SD to generate high-quality outputs.
FaithDiff~\cite{faithdiff} integrates the encoder, alignment module, and diffusion model into a unified trainable framework, enabling end-to-end joint optimization and achieving improved restoration results.
DiT4SR~\cite{dit4sr} further utilizes a pre-trained Diffusion Transformer (DiT) to achieve efficient and scalable super-resolution restoration.
However, a key limitation of these approaches is their tendency to employ a uniform processing strategy that fails to account for the distinct characteristics of different degradation types, often leading to structural distortions or loss of fine details.
To this end, we introduce a new framework that integrates a Mixture-of-Experts (MoE) module into a pretrained diffusion model. 
This integration provides a unified solution for diverse degradations without compromising the generative capacity of the model.

\section{Preliminaries}

\textbf{Diffusion Transformer.}
We build upon the Diffusion Transformer (DiT)~\cite{dit} architecture as implemented in Stable Diffusion 3.5, which adopts a rectified flow-matching formulation~\cite{RectifiedFlow, flowmatching} for training and sampling. 
Unlike traditional diffusion models based on DDPM~\cite{DDPM} that rely on a fixed forward noising process, rectified flow models learn a continuous velocity field that directly transports samples from noise to data along straight-line paths in the data space. 
Formally, given a noise sample $\mathbf{z} \sim \mathcal{N}(0, \mathbf{I})$ and a clean image $\mathbf{x}$, the model learns a velocity function $\mathbf{v}_\theta(\mathbf{x}_t, t)$ :
\begin{equation}
    \frac{d\mathbf{x}_t}{dt} = \mathbf{v}_\theta(\mathbf{x}_t, t),
\end{equation}
where $\mathbf{x}_t = t\mathbf{x} + (1-t)\mathbf{z}$ denotes the interpolation path between noise and data at time $t \in [0,1]$.
The DiT backbone in SD3.5 employs a Transformer-based architecture to predict this velocity field, processing patch-embedded image tokens conditioned on time embeddings and optional context features (e.g., text embeddings). 
This formulation enables more efficient sampling and better preservation of semantic structures, making it particularly suitable for high-fidelity image restoration tasks.
In our work, we leverage the pretrained DiT from SD3.5 as the backbone due to its strong generative priors and ability to model diverse image distributions within a unified framework. 
Unlike task-specific restoration models that are limited to narrow degradation families, DiT is originally trained on large-scale image text datasets and encodes rich semantic and perceptual knowledge. This enables high-quality reconstruction across a wide range of visual content and degradation types
%
%
Furthermore, its Transformer-based architecture provides flexible feature representations that can be effectively adapted to various low-level vision tasks, making it an ideal foundation for our unified restoration framework.
\vspace{-1mm}
{\flushleft\textbf{Mixture-of-Experts.}}
The Mixture-of-Experts (MoE)~\cite{moe} paradigm is a conditional computation framework that decomposes a model into multiple specialized sub-networks known as experts. 
Instead of activating all parameters for every input, MoE dynamically selects a subset of experts based on the input characteristics through a trainable gating mechanism. 
Formally, given an input feature $\mathbf{x}$, the output of an MoE layer is computed as:
\begin{equation}
    \mathbf{y} = \sum_{i=1}^{E} \mathrm{g}_i(\mathbf{x}) \odot f_i(\mathbf{x}),
\end{equation}
where $E$ is the total number of experts, $\odot$ denotes element-wise multiplication, $f_i(\cdot)$ denotes the $i$-th expert network, and $\mathrm{g}_i(\cdot)$ represents the corresponding gating function that determines the contribution of each expert.
Depending on the gating design, some approaches employ sparse MoE, which activates only a small subset of k experts per input via Top-k routing to scale capacity efficiently, while others use dense MoE, which activates every expert for every input and modulates their strengths through the gating weights. 
For all-in-one image restoration, MoE offers a particularly compelling advantage: it enables a single model to dynamically adapt its computational pathways to different degradation types, achieving task-specific specialization without the need for separate models or task-specific fine-tuning. This makes it an ideal framework for handling the diverse and often mixed degradations encountered in real-world scenarios.

\section{Proposed Method}
To effectively handle diverse degradation types with a unified model, we propose a hierarchical MoE in MoE architecture integrated with a pretrained DiT, termed MiM-DiT, which enables dynamic adaptation to different degradation patterns while preserving the favorable generative capacity of the pretrained backbone. 
By incorporating both Inter-MoE and Intra-MoE routing mechanisms, our framework adaptively selects the most suitable experts at both task and feature levels, facilitating the capture of global contextual structures and fine local details according to each degradation type.
Through this dynamic specialization, our approach maintains the high generative quality of the diffusion backbone while adapting to a wide spectrum of degradation patterns, achieving robust all-in-one restoration.
%
\subsection{Hierarchical MoE in MoE}
\label{sec:degradation_aware_moe}
%

The proposed hierarchical Mixture-of-Experts in Mixture-of-Experts (MiM) architecture is designed to capture both structural diversity and granular variations across image degradations. 
It operates at two complementary levels: Inter-MoE and Intra-MoE. 
The Inter-MoE level integrates multiple heterogeneous expert groups, each built upon a distinct attention mechanism, through dense fusion, enabling the model to adapt to fundamentally different degradation types.
The Intra-MoE level further incorporates sparse MoE modules with Top-$k$ routing to handle fine-grained variations within each degradation category. 
This dual-level design explicitly decouples inter-degradation structural adaptation from intra-degradation specialization, allowing the model to jointly address coarse-grained architectural selection and fine-grained parameter modulation, which is difficult to achieve in single-level MoE frameworks.

At the Inter-MoE level, we define four expert groups: (1) \textit{spatial self-attention} effective for modeling long-range dependencies; (2) \textit{channel self-attention}, particularly effective for channel-wise feature recalibration;  (3) \textit{Swin attention}, which balances local and global context with efficiency; and (4) \textit{SE attention}, ideal for global illumination modeling in tasks such as dehazing and low-light enhancement. 
We employ a dense router, implemented as a lightweight gating network, to fuse the outputs of all experts. This router predicts adaptive fusion weights $\mathrm{g}(\mathbf{x})$ from the features of the low-quality input. The final output is computed as:
\vspace{-2mm}
\begin{equation}
    \mathbf{y} = \sum_{i=1}^{4} \mathrm{g}^{dense}_i(\mathbf{x_{(inter)}}) \odot \mathcal{F}_i(\mathbf{x_{(inter)}}),
\end{equation}
where $\mathcal{F}_i(\cdot)$ denotes the output of the $i$-th expert group. This dense fusion allows the model to simultaneously leverage complementary inductive biases. 
For example, spatial attention captures long-range dependencies while Swin attention enhances local context, resulting in more comprehensive restoration.

Within each expert group $\mathcal{F}_i$, we further introduce an Intra-MoE layer, implemented as a sparse router based on a Top-$k$ gating mechanism, where $\mathcal{F}_i(\mathbf{x})$ is computed as:
\vspace{-1mm}
\begin{equation}
    \mathcal{F}_i(\mathbf{x}) = \sum_{j \in \mathcal{S}_i} \mathrm{g}^{sparse}_j(\mathbf{x_{intra}}) \odot f_{i,j}(\mathbf{x_{intra}}),
\end{equation}
\vspace{-1mm}
where the Intra-MoE module consists of $N$ sub-experts, all of which share the same attention architecture but have independently trained parameters. 
For each input, a Top $k$ gating mechanism selects the most relevant sub-experts with $\mathcal{S}_i$ being the set of activated sub-experts $g_j(\cdot)$ representing gating scores and $f_{i,j}(\cdot)$ denoting individual sub-expert networks. 
This enables detailed adaptation to variations within a degradation type, such as short versus long exposure in motion blur or light versus heavy haze in dehazing.
The gating networks at both levels utilize linearly projected input features to enable routing decisions without external labels. 
The Inter-MoE dense router directly predicts fusion weights, while the Intra-MoE sparse router first applies a softmax normalization followed by top-$k$ selection to activate the most relevant sub-experts.
By combining dense structural fusion with sparse intra-structure specialization, our MoE-in-MoE architecture achieves both broad generalization and fine-grained adaptability. 
It fully leverages the strong generative priors of the pretrained DiT backbone while dynamically adjusting to diverse and mixed degradation patterns in a unified, efficient manner.
We provide more details in the supplementary materials.

\begin{table*}[!t]\fontsize{6pt}{5.5pt}\selectfont
\centering
\setlength{\tabcolsep}{3pt}

\vspace{-4mm}

    \caption{Quantitative evaluations of the proposed method against state-of-the-art ones on FoundIR~\cite{foundir} benchmarks. Models marked with an asterisk ($^*$) are evaluated using officially released pre-trained weights, while those without are retrained following the FoundIR~\cite{foundir} on the same dataset for fair comparison. The best and second performances are marked in \textcolor{red}{red} and \textcolor{blue}{blue}, respectively.
    }
    \label{tab:foundir_dataset}
\vspace{-3mm}
\begin{tabular}{l|l|ccccccccccccc}
        \toprule
Degradation                      & Metrics           & AirNet~\cite{airnet}   &  DGUNet~\cite{dgunet}  &TransWeather~\cite{transweather} &PromptIR~\cite{promptir} & DiffIR~\cite{diffir}  & DiffUIR~\cite{DiffUIR} & DA-CLIP~\cite{DA-CLIP}  & AutoDIR*~\cite{autodir}  & FoundIR~\cite{foundir} &DiT4SR~\cite{dit4sr}& Ours   \\
        \midrule 

\multirow{7}{*}{Blur}        
                             & LPIPS~$\downarrow$    &0.4340    &0.3855    &0.3911       &0.3900   &0.3692   &\textcolor{blue}{0.2043}   &0.3067   &0.3291   &\textcolor{red}{0.1722}  & 0.3019    &0.2372  \\
                             & FID~$\downarrow$      &42.4248   &35.3296   &34.3489      &36.2908  &33.8067  &\textcolor{red}{9.0676}   &16.9121  &17.8850  &\textcolor{blue}{10.5639} & 29.0668   &16.9659  \\
                             & NIQE~$\downarrow$     &7.6429    &7.7016    &7.2645       &7.8885   &7.2844   &5.5757   &6.4046   &6.6792   &5.5645  &\textcolor{blue}{2.9719}   & \textcolor{red}{2.9546}   \\
                             & LIQE~$\uparrow$       &1.0102    &1.0089    &1.0042       &1.0086   &1.0127   &1.8270   &1.0274   &1.0140   &2.0964  &\textcolor{blue}{3.0184}  &\textcolor{red}{3.9440}\\
                             & MUSIQ~$\uparrow$      &22.9102   &22.8828   &21.1029      &23.1877  &23.5724  &50.8270  &31.3108  &30.8041  & 53.4637&\textcolor{blue}{66.8670}   & \textcolor{red}{68.5028}   \\
                             & CLIP-IQA~$\uparrow$   &0.3677    &0.3199    &0.3121       &0.2929   &0.2981   &0.4806   &0.4772   &0.3789   &0.5083  &\textcolor{blue}{0.6242}   & \textcolor{red}{0.6405}      \\
                                     \midrule 
\multirow{7}{*}{Noise}       
                             & LPIPS~$\downarrow$    &0.3419    &0.1660    &0.2942       &0.1722   &\textcolor{blue}{0.1109}   &0.1200   &0.2733    &0.1745   &0.1016  &0.1145   &\textcolor{red}{0.0660}    \\
                             & FID~$\downarrow$      &33.3680   &5.9338    &15.9013      &5.7243   &5.6000   &5.6548   &11.7462   &\textcolor{blue}{4.9221}   &\textcolor{red}{3.5635}  &29.6721   &{ 11.7939}     \\

                             & NIQE~$\downarrow$     &7.4690    &7.3793    &7.4903       &7.5749   &\textcolor{red}{5.0059}   &5.4719   &7.5410    &5.3673   &5.7275  &\textcolor{blue}{5.3522}   &5.8004  \\
                             & LIQE~$\uparrow$                  &1.0297    &1.1668    &1.0246      &1.1814   &1.7451   &1.8750   &1.0305   &1.9007   &1.9093  &\textcolor{blue}{2.1045} &\textcolor{red}{2.4710}\\
                             & MUSIQ~$\uparrow$      &31.4204   &38.604    &28.2514      &38.1951  &44.8197  &46.0562  &30.4246   &44.6805  &45.3634 &\textcolor{blue}{47.2216}   &\textcolor{red}{48.1446}  \\
                             & CLIP-IQA~$\uparrow$   &0.5035    &0.5182    &0.4182       &0.5180   &0.5349   &0.5314   &0.5082    &0.5165   &\textcolor{red}{0.5672}  &0.5412   &\textcolor{blue}{0.5484}  \\
                                     \midrule 
\multirow{7}{*}{Haze}        
                             & LPIPS~$\downarrow$    &0.6294    &0.4600    &0.5780       &0.4627   &0.2310   &0.2071   &0.6079    &0.4026   &\textcolor{blue}{0.1918}  &0.3448   &\textcolor{red}{0.1832}    \\
                             & FID~$\downarrow$      &77.6472   &59.0215   &45.7699      &45.2893  &46.7080  &\textcolor{red}{23.9144}  &49.9409  &44.0517  &\textcolor{blue}{29.9501} &65.2104   &30.5465    \\

                             & NIQE~$\downarrow$     &6.9454    &5.6843    &6.9400       &5.6525   &3.1873   &3.3850   &7.0078    &4.7216   &3.6488  &\textcolor{blue}{3.2142}   &\textcolor{red}{2.9048}\\
                             & LIQE~$\uparrow$                  &1.0386    &1.3701    &1.0118       &1.4435   &2.3967   &2.8992   &1.0335   &1.4481   &3.1309  &\textcolor{blue}{3.3618} &\textcolor{red}{3.6647}\\
                             & MUSIQ~$\uparrow$      &32.0978   &39.1956   &27.6606      &40.2472  &57.2363  &57.7686  &31.0622   &43.5787  &\textcolor{blue}{59.8311} &57.7600   &\textcolor{red}{62.9768}\\
                             & CLIP-IQA~$\uparrow$   &0.4970    &0.5105    &0.3695       &0.5045   &0.5263   &\textcolor{blue}{0.5741}   &0.4959    &0.4878   &0.5709  &0.5297   &\textcolor{red}{0.6276}\\

                                     \midrule 
\multirow{7}{*}{Rain}        
                             & LPIPS~$\downarrow$    &0.4855    &0.2635    &0.5118       &0.2500   &0.2574   &\textcolor{blue}{0.1097}   &0.4347    &0.3684   &\textcolor{red}{0.0795}  &0.1570   &0.1142  \\
                             & FID~$\downarrow$      &88.7306   &57.5912   &70.9293      &51.5580  &76.9926  &\textcolor{blue}{15.3162}  &83.8305   &40.9156  &\textcolor{red}{11.8318} &27.5860   &19.0445 \\

                             & NIQE~$\downarrow$     &7.7079    &6.2807    &8.1970       &6.3012   &5.1983   &4.5472   &7.7218    &7.3112   &4.5854  &\textcolor{blue}{4.5005}   &\textcolor{red}{4.4349} \\
                             & LIQE~$\uparrow$                  &1.0756    &1.1202    &1.0618       &1.1860   &1.3513   &1.3018   &1.0927   &1.0416   &1.4309  &1.3248 &\textcolor{blue}{1.3585}\\
                             & MUSIQ~$\uparrow$      &23.4481   &27.2026   &20.2728      &27.6832  &30.9400  &\textcolor{blue}{30.9973}  &23.7416   &23.5576  &30.8493 &30.4608   &\textcolor{red}{31.0638} \\
                             & CLIP-IQA~$\uparrow$   &0.5362    &0.5140    &0.4150       &0.5070   &0.4478   &0.4563   &\textcolor{blue}{0.5509}    &0.4782   &0.4599  &\textcolor{red}{0.5525}   &{0.4546} \\
                                     \midrule 

\multirow{7}{*}{Low-light}    
                             & LPIPS~$\downarrow$    &0.8002    &0.4795    &0.5747       &0.4873   &0.3392   &0.3071   &0.5067    &0.3563   &\textcolor{blue}{0.2739}  &0.6151   &\textcolor{red}{0.2176} \\
                             & FID~$\downarrow$      &185.2157  &91.7658   &119.2019     &90.6817   &63.7529  &\textcolor{blue}{51.3239}  &119.9456  &\textcolor{red}{42.0912}  &54.6338 &160.1224   &56.0566  \\

                             & NIQE~$\downarrow$     &10.6218   &7.8858    &7.8279       &7.8012   &\textcolor{red}{4.6574}   &5.6757   &7.9703    &7.6149   &6.4992  &7.1909   &\textcolor{blue}{5.1561} \\  
                             & LIQE~$\uparrow$                  &1.0504    &1.0274    &1.0211       &1.0393   &1.3458   &1.9825   &1.0239   &1.0724   &1.9616  &\textcolor{blue}{2.4486} &\textcolor{red}{3.1281}\\

                             & MUSIQ~$\uparrow$      &20.6885   &28.7058   &22.3165      &27.6250  &42.8427  &47.6626  &25.7504   &31.5460  &\textcolor{blue}{49.0025} &29.6363   &\textcolor{red}{56.7996} \\
                             & CLIP-IQA~$\uparrow$   &0.3991    &0.3893    &0.3412       &0.3837   &0.4549   &0.5500   &0.4169    &0.4753   &0.5125  &\textcolor{blue}{0.5578}   &\textcolor{red}{0.6104} \\
                                     \bottomrule

\end{tabular}

\end{table*}

\begin{table*}[!t]\fontsize{6pt}{5.5pt}\selectfont
\centering
\setlength{\tabcolsep}{3pt}
\vspace{-3mm}

    \caption{Quantitative evaluations of the proposed method against state-of-the-art ones on additional public benchmarks, including 4KRD~\cite{4krd}, RealRain-1K~\cite{realrain}, HazeRD~\cite{hazerd}, and UHD-LL~\cite{uhdll}. Models marked with an asterisk ($^*$) are evaluated using officially released pre-trained weights, while those without are retrained following the FoundIR~\cite{foundir} on the same dataset for fair comparison. The best and second performances are marked in \textcolor{red}{red} and \textcolor{blue}{blue}, respectively.}
    \label{tab: public_benchmark}
\vspace{-3mm}
\begin{tabular}{l|l|ccccccccccccc}
        \toprule
Dataset & Metrics & TransWeather~\cite{transweather}  & AirNet~\cite{airnet}&  PromptIR~\cite{promptir} & PromptIR* & DiffIR~\cite{diffir} & DiffUIR~\cite{DiffUIR} & DiffUIR*& AutoDIR*~\cite{autodir} & FoundIR~\cite{foundir} & DiT4SR~\cite{dit4sr} & Ours \\
        \midrule

\multirow{7}{*}{4KRD~\cite{4krd}} 
                      & LPIPS~$\downarrow$ &0.4468  &0.4837    &0.4421  &0.3130  &\textcolor{blue}{0.2373}  &0.2669   &0.2421  &0.3328  &\textcolor{red}{0.2347}  &0.3283 &0.2896 \\
                      & FID~$\downarrow$   &27.5878 &32.6022   &25.6506 &34.3458 &\textcolor{red}{14.6920} &21.1369  &15.5659 &\textcolor{blue}{15.2305} &16.6398 &23.4762 &18.1328 \\
                      & NIQE~$\downarrow$  &7.9220  &8.1894    &8.8772  &6.1589  &6.4251  &6.0133   &5.7992  &7.3755  &6.0495  &\textcolor{blue}{4.5789} &\textcolor{red}{3.6536} \\
                      & LIQE~$\uparrow$   &1.0119   &1.0115   &1.0192   &1.0531   &1.1162   &1.0711   &1.0795   &1.0413   &1.1014   &\textcolor{blue}{2.2047}   &\textcolor{red}{2.5894}    \\
                      & MUSIQ~$\uparrow$   &17.6972 &18.7134   &18.9286 &23.9315 &27.6394 &26.1672  &28.0036 &22.1480 &27.3308 &\textcolor{blue}{40.8834} &\textcolor{red}{42.0540} \\
                      & CLIP-IQA~$\uparrow$&0.3305  &0.4515    &0.4172  &0.3531  &0.3895  &0.3849   &0.3842  &0.3516  &0.4010  &\textcolor{blue}{0.5136} &\textcolor{red}{0.5587} \\
        \midrule
\multirow{7}{*}{HazeRD~\cite{hazerd}} 
                      & LPIPS~$\downarrow$ &0.4064  &0.4363   &0.4243  &0.2619  &0.2651  &\textcolor{blue}{0.2559}  &0.2705  &0.2881  &\textcolor{red}{0.1918}   &0.2883 &0.2657 \\
                      & FID~$\downarrow$   &57.0182 &93.7266  &49.2953 &94.5781 &69.2957 &62.3962 &72.9453 &\textcolor{blue}{45.2901} &56.6236  &45.3254 &\textcolor{red}{44.4432} \\
                      & NIQE~$\downarrow$  &7.4374  &7.4258   &7.9159  &\textcolor{red}{4.2298}  &\textcolor{blue}{4.4926}  &5.7842  &5.7349  &6.8089  &5.4319   &4.8517 &4.9580 \\
                      & LIQE~$\uparrow$   &1.0862   &1.0607   &1.0995   &1.3626   &1.3324   &1.4233   &1.3435   &1.2380   &1.2689   &\textcolor{blue}{1.4538}   &\textcolor{red}{1.8930}    \\
                      & MUSIQ~$\uparrow$   &20.3614 &23.5342  &21.6588 &31.8317 &27.5206 &30.7580 &29.6138 &26.4155 &29.2933  &\textcolor{blue}{33.8382} &\textcolor{red}{34.9328} \\
                      & CLIP-IQA~$\uparrow$&0.3782  &0.4928   &0.4542  &0.4296  &0.4395  &0.4819  &0.4696  &0.4445  &0.4598   &\textcolor{blue}{0.5016} &\textcolor{red}{0.5208} \\
        \midrule

\multirow{7}{*}{RealRain-1K~\cite{realrain}}
                      & LPIPS~$\downarrow$ &0.3630  &0.4324    &0.3544   &0.4490   &0.4901   &\textcolor{blue}{0.2551}   &0.4610   &0.4710   &\textcolor{red}{0.1345}  &0.4015 &0.3912 \\
                      & FID~$\downarrow$   &152.5585&172.8349  &157.8070 &164.4184 &162.8672 &\textcolor{blue}{125.1769} &167.6739 &170.1762 &\textcolor{red}{92.5971} &151.8471 &{150.0357} \\
                      & NIQE~$\downarrow$  &7.9606  &\textcolor{blue}{7.1477}    &8.2378   &6.4019   &9.8762   &8.1173   &7.7364   &7.6838   &9.9191  &7.5494 &\textcolor{red}{6.7482} \\
                      & LIQE~$\uparrow$   &1.1777   &1.1998   &1.1713   &1.2399   &1.1484   &1.2394   &1.2619   &1.1841   &1.2385   &\textcolor{blue}{1.3947}   &\textcolor{red}{1.4782}    \\
                      & MUSIQ~$\uparrow$   &31.3593 &34.5045   &35.4560  &35.8707  &36.3306  &34.4170  &\textcolor{blue}{38.6514}  &37.7553  &29.5940 &38.1462 &\textcolor{red}{39.9971} \\
                      & CLIP-IQA~$\uparrow$&0.3562  &\textcolor{red}{0.3926}    &0.3719   &0.3562   &0.3525   &0.3435   &0.3563   &0.3644   &0.3614  &0.3713 &\textcolor{blue}{0.3802} \\
        \midrule

\multirow{7}{*}{UHD-LL~\cite{uhdll}} 
                      & LPIPS~$\downarrow$ &0.5002  &0.6043    &0.4878  &0.5886  &0.4627  &0.3970  &0.5804  &\textcolor{blue}{0.3843}  &\textcolor{red}{0.3214}  &0.4638 &{0.4061} \\
                      & FID~$\downarrow$   &61.5787 &84.7410   &55.1285 &74.9124 &59.6029 &\textcolor{blue}{51.4430} &69.6643 &\textcolor{red}{36.8954} &56.6126 &76.9471 &{71.6686} \\
                      & NIQE~$\downarrow$  &8.2414  &9.4197    &9.4333  &5.1378  &\textcolor{red}{4.4254}  &5.4931  &5.2791  &6.3458  &6.3828  &4.8384 &\textcolor{blue}{4.6243} \\
                      & LIQE~$\uparrow$   &1.0258   &1.028   &1.0419   &1.1841   &1.1814   &1.309   &1.1747   &1.1787   &1.4112   &\textcolor{blue}{1.6474}   &\textcolor{red}{1.9332}    \\
                      & MUSIQ~$\uparrow$   &22.602  &23.3077   &23.6453 &26.7194 &34.1731 &33.7250 &27.1415 &28.0119 &39.2011 &\textcolor{blue}{39.5840} &\textcolor{red}{40.6206} \\
                      & CLIP-IQA~$\uparrow$&0.3417  &0.4095    &0.4195  &0.4167  &0.4287  &0.4558  &0.4040  &0.4212  &0.4791  &\textcolor{blue}{0.5043} &\textcolor{red}{0.5214} \\
        \bottomrule

\end{tabular}
\vspace{-5mm}
\end{table*}

\subsection{Conditional Generation for DiT with MiM}
\label{sec:conditioning}
Building upon the hierarchical MoE architecture defined in Section~\ref{sec:degradation_aware_moe}, we employ the MiM module as a powerful conditioning mechanism for the DiT backbone. 
The purpose of this integration is to transform the low-quality input into a feature representation that carries a high-quality restoration prior, thereby providing effective guidance throughout the generative reconstruction process of the Diffusion Transformer.
Specifically, the latent LQ image is first processed through MiM, which produces a feature map that captures clear structural priors.
This feature map is then transformed by a Zero Linear layer (a linear projection initialized with zero weights).
The resulting conditioning features are concatenated with the current latent token sequence, forming an augmented input to the attention layer.
For the $n$-th block in the proposed MiM-DiT, the detailed computation proceeds as follows:
\begin{equation}
\begin{split}
&  \mathbf{z}_{\text{cond}}^{(n)} = 
\begin{cases}
  MiM^{(n)}(\mathbf{z}_{\text{LQ}}) & \text{if } n = 0, \\
 MiM^{(n)}(\mathbf{z}_{\text{MiM}}^{(n)}) & \text{if } n \geq 1. \\
\end{cases}\\
&\mathbf{z}_{\text{cat}}^{(n)} = Concat(\mathbf{z}_{\text{Text}}^{(n)};\mathbf{z}_{\text{DiT}}^{(n)}; ZeroLinear(\mathbf{z}_{\text{cond}}^{(n)})),\\
&\mathbf{z}_{\text{out}}^{(n)} = MLP^{(n)}(Attention^{(n)}(\mathbf{z}_{\text{cat}}^{(n)})),\\
&\mathbf{z}_{\text{Text}}^{(n+1)};\mathbf{z}_{\text{DiT}}^{(n+1)}; \mathbf{z}_{\text{MiM}}^{(n+1)} = Split(\mathbf{z}_{\text{out}}^{(n)}),
\end{split}
\end{equation}
%
%
where $\mathbf{z}_{\text{LQ}}$ denotes the latent representation of the LQ image, $\mathbf{z}_{\text{Text}}^{(n)}$ denotes the text representation to the $n$-th block, $\mathbf{z}_{\text{DiT}}^{(n)}$ denotes the input latent features to the $n$-th block: noise input when $n = 0$, and the output from the previous MiM-DiT block when $n \geq 1$, $MiM^{(n)}$ denotes the proposed MiM module in the $n$-th block, $MLP^{(n)}$ and $Attention^{(n)}$ denote the self-attention and MLP operations in the $n$-th block, respectively, $ZeroLinear$ denotes a linear projection layer with zero initialization, $Concat$ denotes the operation of concatenation in the token length dimension, and $Split$ denotes the operation for splitting the input token along the token length dimension into two parts.
We use $\mathbf{z}_{\text{DiT}}$ in the last block as the output $\mathbf{Z}_t$ of the diffusion process at the time step $t$. After iterating over $T$ steps, we decode the resulting clean latent variable $\mathbf{Z}_0$, and finally reconstruct the restored image through the VAE decoder.
Benefiting from the zero-initialized linear transformation applied to the features output by MiM, which serve as the conditioning signal guiding the pre-trained DiT, this design ensures minimal influence of the conditioning signal at the beginning of training, thereby effectively maintaining optimization stability. 
During training, the model gradually learns to leverage this conditioning signal to achieve high-quality image restoration.
\begin{figure*}[!t]
\footnotesize
\centering
    \begin{tabular}{c c c c c c c}
            \multicolumn{3}{c}{\multirow{5}*[35.5pt]{
            \hspace{-3mm} 
            \includegraphics[width=0.325\linewidth,height=0.195\linewidth]{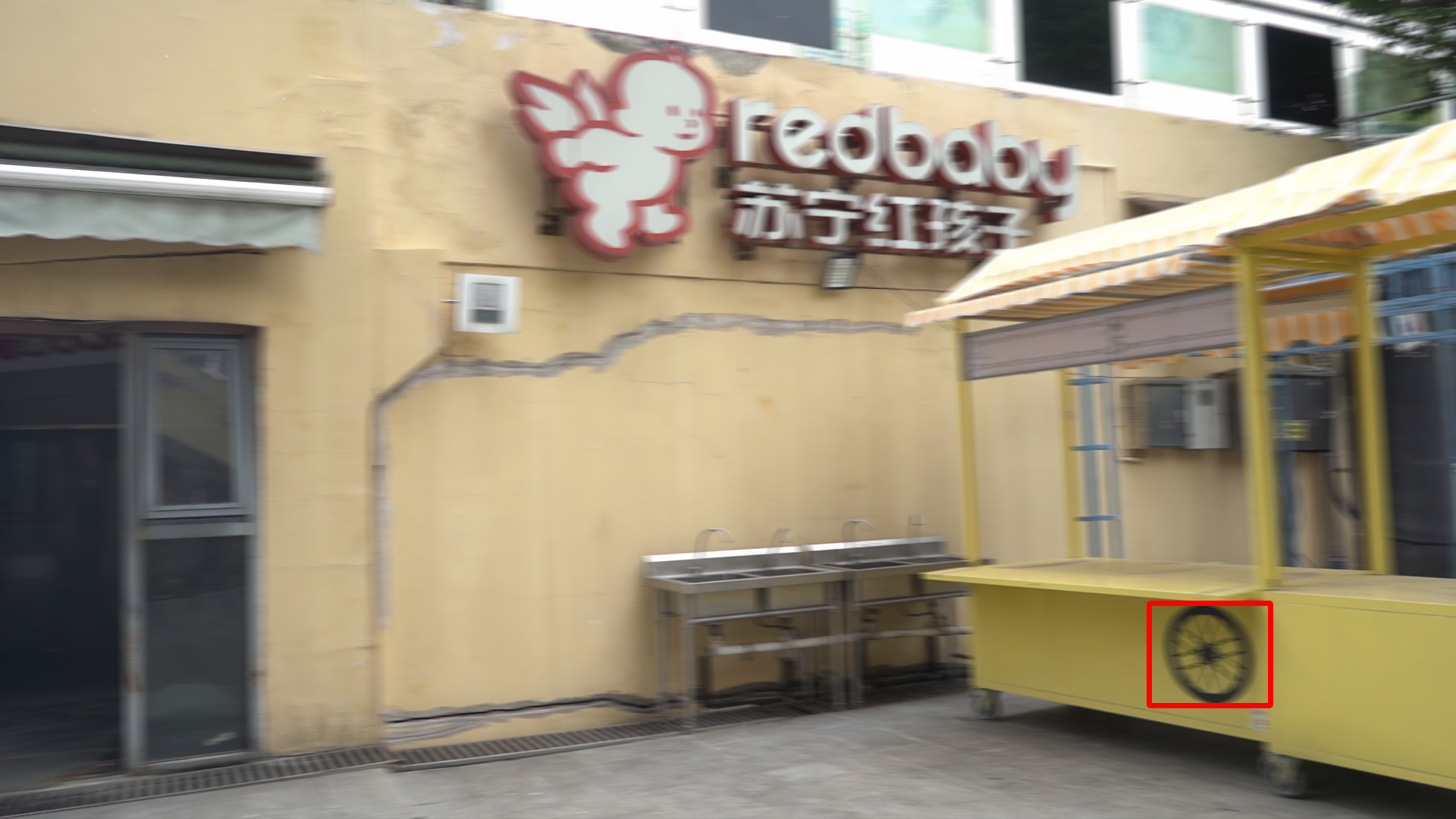}}}
            & \hspace{-4.0mm} \includegraphics[width=0.16\linewidth,height=0.085\linewidth]{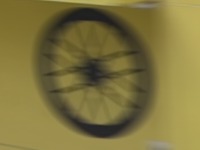}
            & \hspace{-4.0mm} \includegraphics[width=0.16\linewidth,height=0.085\linewidth]{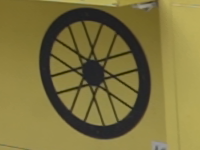}
            & \hspace{-4.0mm} \includegraphics[width=0.16\linewidth,height=0.085\linewidth]{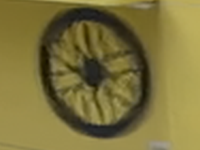}
            & \hspace{-4.0mm} \includegraphics[width=0.16\linewidth,height=0.085\linewidth]{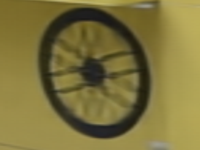}
              \\
    		\multicolumn{3}{c}{~}
            & \hspace{-4.0mm} (a) Blurred patch
            & \hspace{-4.0mm} (b) GT
            & \hspace{-4.0mm} (c) DA-CLIP~\cite{DA-CLIP}
            & \hspace{-4.0mm} (d) AutoDIR~\cite{autodir} \\		
    	\multicolumn{3}{c}{~}

            & \hspace{-4.0mm} \includegraphics[width=0.16\linewidth,height=0.085\linewidth]{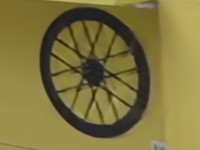}
            & \hspace{-4.0mm} \includegraphics[width=0.16\linewidth,height=0.085\linewidth]{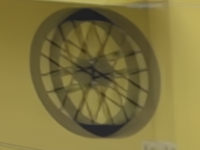}
            & \hspace{-4.0mm} \includegraphics[width=0.16\linewidth,height=0.085\linewidth]{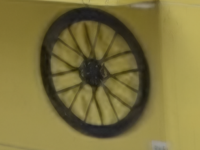}
            & \hspace{-4.0mm} \includegraphics[width=0.16\linewidth,height=0.085\linewidth]{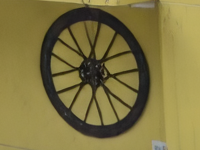}
            \\
    	\multicolumn{3}{c}{\hspace{-4.0mm} Blurred image}
            & \hspace{-4.0mm} (e) DiffUIR~\cite{DiffUIR}
            & \hspace{-4.0mm} (f) FoundIR~\cite{foundir}
            & \hspace{-4.0mm} (g) DiT4SR~\cite{dit4sr}
            & \hspace{-4.0mm} (h) Ours\\

    \end{tabular}
\vspace{-4mm}

\caption{Deblurred results on the FoundIR dataset~\cite{foundir}. The deblurred results in (c)-(g) still contain significant blur effects. In contrast, our method generates clear results.}
\label{fig: result_1}
\vspace{-2mm}
\end{figure*}

\begin{figure*}[!t]
\footnotesize
\vspace{-2mm}
\centering
    \begin{tabular}{c c c c c c c}
            \multicolumn{3}{c}{\multirow{5}*[35.5pt]{
            \hspace{-3mm} 
            \includegraphics[width=0.325\linewidth,height=0.195\linewidth]{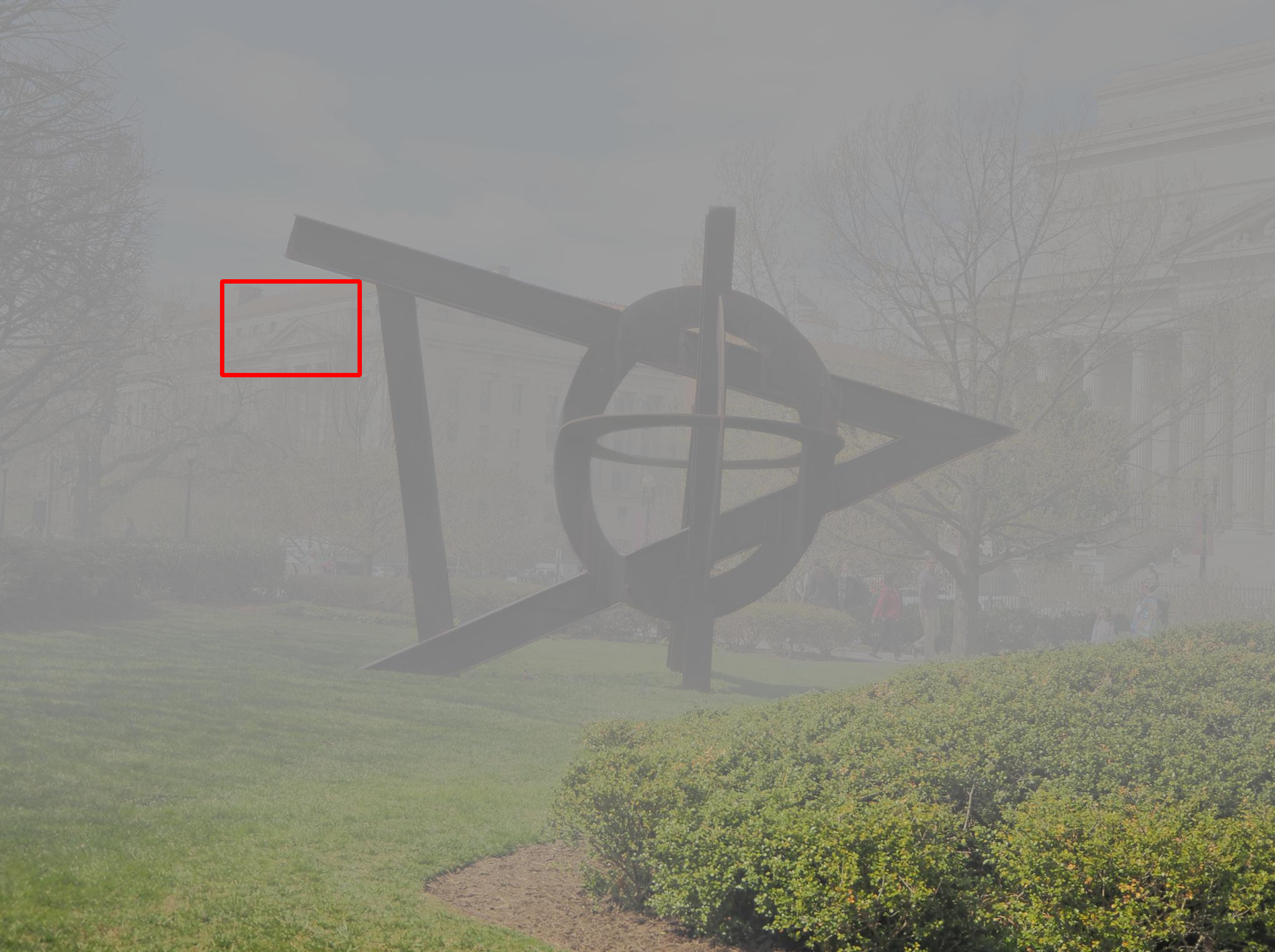}}}
            & \hspace{-4.0mm} \includegraphics[width=0.16\linewidth,height=0.085\linewidth]{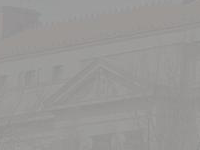}
            & \hspace{-4.0mm} \includegraphics[width=0.16\linewidth,height=0.085\linewidth]{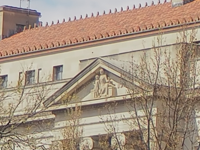}
            & \hspace{-4.0mm} \includegraphics[width=0.16\linewidth,height=0.085\linewidth]{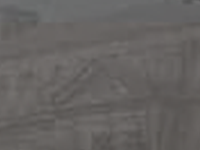}
            & \hspace{-4.0mm} \includegraphics[width=0.16\linewidth,height=0.085\linewidth]{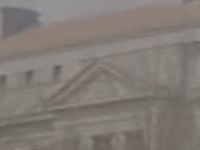}
              \\
    		\multicolumn{3}{c}{~}
            & \hspace{-4.0mm} (a) Hazy patch
            & \hspace{-4.0mm} (b) GT
            & \hspace{-4.0mm} (c) DA-CLIP~\cite{DA-CLIP}
            & \hspace{-4.0mm} (d) AutoDIR~\cite{autodir} \\		
    	\multicolumn{3}{c}{~}
            & \hspace{-4.0mm} \includegraphics[width=0.16\linewidth,height=0.085\linewidth]{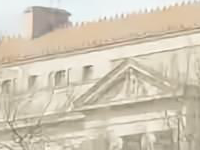}
            & \hspace{-4.0mm} \includegraphics[width=0.16\linewidth,height=0.085\linewidth]{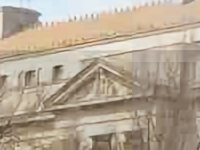}
            & \hspace{-4.0mm} \includegraphics[width=0.16\linewidth,height=0.085\linewidth]{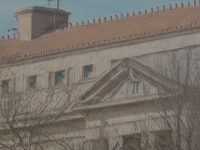}
            & \hspace{-4.0mm} \includegraphics[width=0.16\linewidth,height=0.085\linewidth]{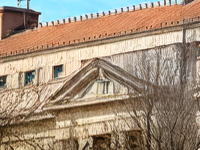}
            \\
    	\multicolumn{3}{c}{\hspace{-4.0mm} Hazy image}
            & \hspace{-4.0mm} (e) DiffUIR~\cite{DiffUIR}
            & \hspace{-4.0mm} (f) FoundIR~\cite{foundir}
            & \hspace{-4.0mm} (g) DiT4SR~\cite{dit4sr}
            & \hspace{-4.0mm} (h) Ours\\

    \end{tabular}
\vspace{-4mm}

\caption{Dehazed results on the FoundIR dataset~\cite{foundir}. The results in (c) to (g) fail to fully restore the original scene content while removing haze. In contrast, our method generates clear and faithful reconstructions.}
\label{fig: result_2}
\vspace{-4mm}
\end{figure*}

\section{Experiments}
In this section, we first describe the datasets and implementation details of the proposed method. 
Then we evaluate our approach against state-of-the-art ones using publicly available benchmark datasets. More experimental results are included in the supplemental material.
\subsection{Experimental Settings}
{\flushleft \textbf{Training datasets.}} 
We train our model on the FoundIR dataset~\cite{foundir}, which provides a large-scale collection of high-quality image pairs captured under controlled conditions. 
FoundIR contains diverse real-world degradation types—including motion blur, haze, rain, low-light, and noise—paired with clean references, making it particularly suitable for training diffusion-based models that require substantial high-fidelity data for stable convergence and high-quality generation. 
The scale and diversity of FoundIR enable our model to learn robust degradation representations and generalize well across various restoration tasks.

{\flushleft \textbf{Evaluation datasets.} }
We evaluate our method on the official test set of FoundIR, as well as four public real-world benchmarks: 4KRD~\cite{4krd} (motion deblurring), RealRain-1K~\cite{realrain} (deraining), HazeRD~\cite{hazerd} (dehazing), and UHD-LL~\cite{uhdll} (low-light enhancement and denoising). 
These datasets contain challenging degraded images, allowing us to assess the generalization and robustness of our method across different degradation types and resolutions. 
Notably, none of these evaluation datasets are used during training, ensuring a fair comparison and demonstrating the out-of-distribution generalization capability of our approach.

{\flushleft \textbf{Implementation details.}} 
We use SD3.5 as the base model.
We employ the Adam optimizer~\cite{Adam} for $2\times10^6$ iterations of training. 
The patch size is empirically set to 512 × 512 pixels, and the batch size is set to 128.
The learning rate is set to a fixed value of \(5 \times 10^{-5}\). 
During inference, we use 40 sampling steps for all tasks.

{\flushleft \textbf{Evaluation metrics.}} 
We employ a range of reference-based and no-reference metrics to evaluate the fidelity and perceptual quality of restored images.
For fidelity and perceptual quality assessment, we employ reference-based metrics, including LPIPS~\cite{lpips} and FID~\cite{FID}.
For no-reference evaluation, we include NIQE~\cite{NIQE}, LIQE~\cite{liqe}, MUSIQ~\cite{MUSIQ}, and CLIPIQA~\cite{CLIP-IQA}, which assess image quality based on statistical and learning-based approaches. 
This diverse set of metrics ensures a thorough analysis of both fidelity and perceptual quality.

\begin{figure*}[!t]
\footnotesize
\centering
    \begin{tabular}{c c c c c c c}
            \multicolumn{3}{c}{\multirow{5}*[35.5pt]{
            \hspace{-3mm} 
            \includegraphics[width=0.325\linewidth,height=0.195\linewidth]{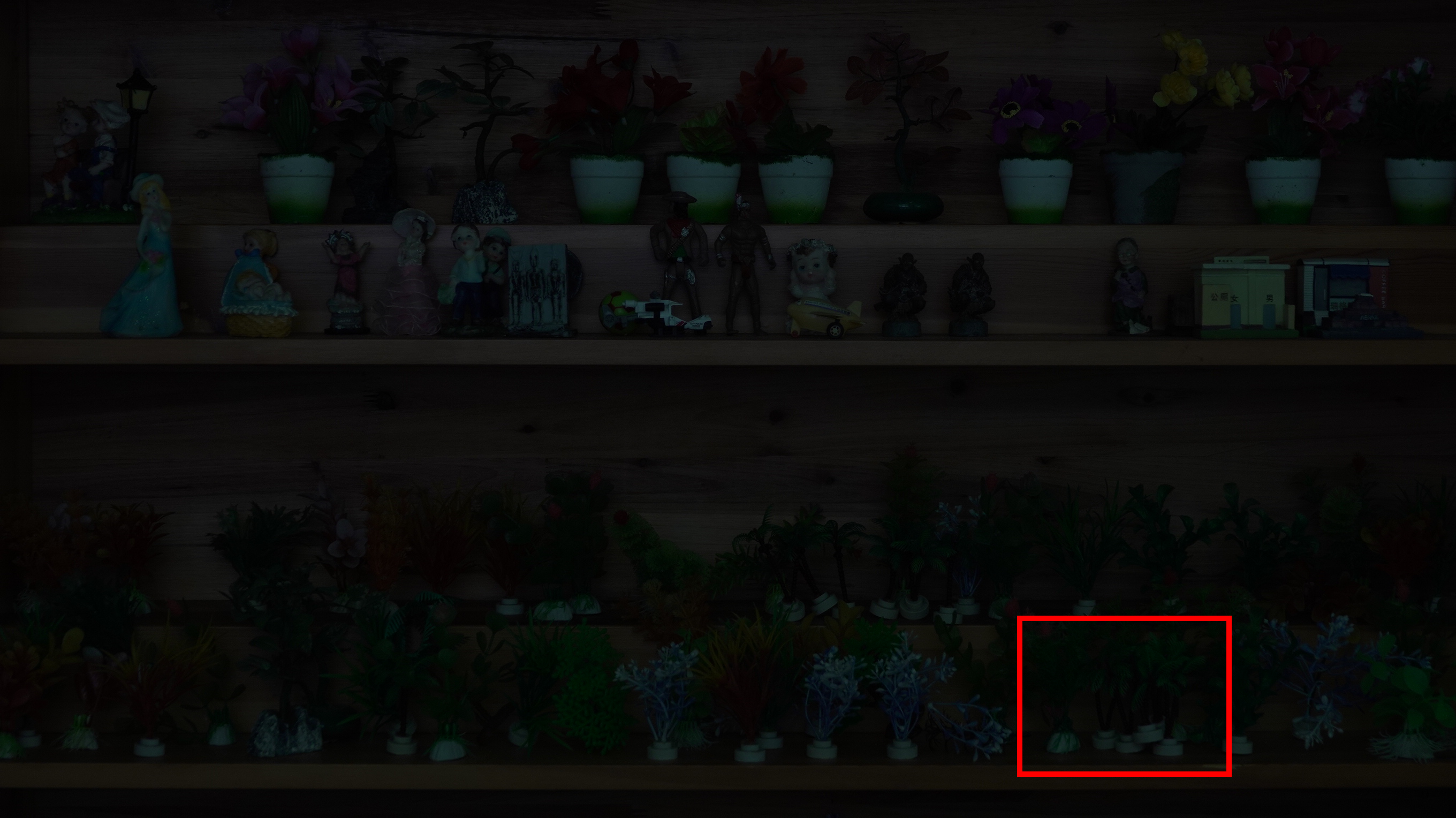}}}
            & \hspace{-4.0mm} \includegraphics[width=0.16\linewidth,height=0.085\linewidth]{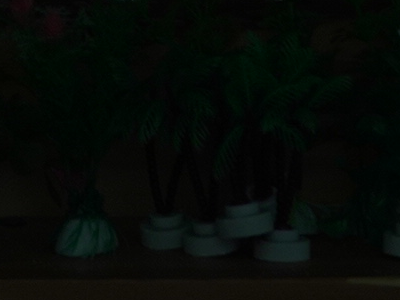}
            & \hspace{-4.0mm} \includegraphics[width=0.16\linewidth,height=0.085\linewidth]{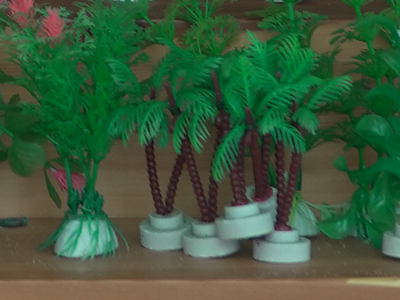}
            & \hspace{-4.0mm} \includegraphics[width=0.16\linewidth,height=0.085\linewidth]{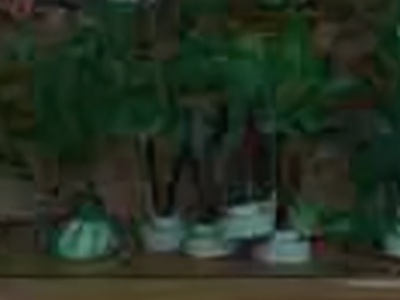}
            & \hspace{-4.0mm} \includegraphics[width=0.16\linewidth,height=0.085\linewidth]{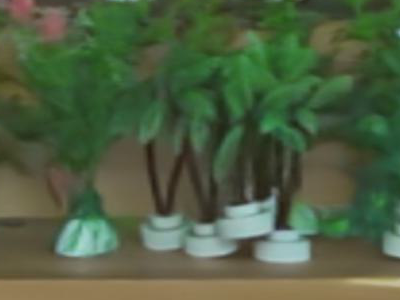}
              \\
    		\multicolumn{3}{c}{~}
            & \hspace{-4.0mm} (a) Low-light patch
            & \hspace{-4.0mm} (b) GT
            & \hspace{-4.0mm} (c) DA-CLIP~\cite{DA-CLIP}
            & \hspace{-4.0mm} (d) AutoDIR~\cite{autodir} \\			
    	\multicolumn{3}{c}{~}
            & \hspace{-4.0mm} \includegraphics[width=0.16\linewidth,height=0.085\linewidth]{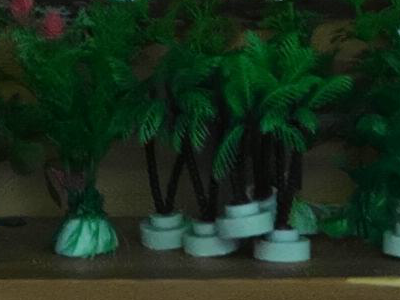}
            & \hspace{-4.0mm} \includegraphics[width=0.16\linewidth,height=0.085\linewidth]{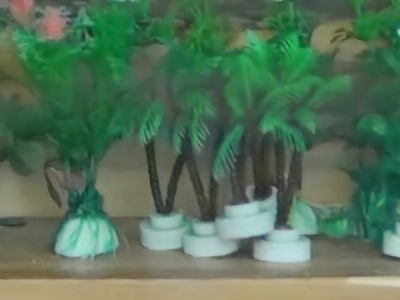}
            & \hspace{-4.0mm} \includegraphics[width=0.16\linewidth,height=0.085\linewidth]{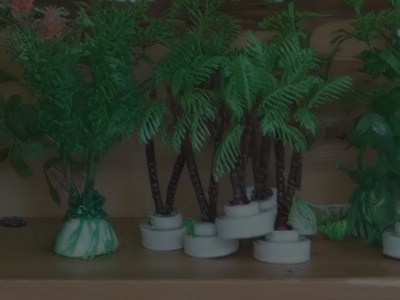}
            & \hspace{-4.0mm} \includegraphics[width=0.16\linewidth,height=0.085\linewidth]{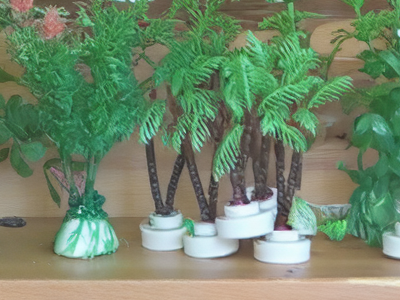}
            \\ 
    	\multicolumn{3}{c}{\hspace{-4.0mm} Low-light image}
            & \hspace{-4.0mm} (e) DiffUIR~\cite{DiffUIR}
            & \hspace{-4.0mm} (f) FoundIR~\cite{foundir}
            & \hspace{-4.0mm} (g) DiT4SR~\cite{dit4sr}
            & \hspace{-4.0mm} (h) Ours\\

    \end{tabular}
\vspace{-4mm}

\caption{Low-light enhanced results on the FoundIR dataset~\cite{foundir}. Results from (c) to (g) suffer from color casts and detail smearing. In contrast, our method recovers accurate colors and fine structures.}
\label{fig: result_3}
\vspace{-6mm}
\end{figure*}

\subsection{Comparisons with Other Methods}
We compare our proposed MiM-DiT with several state-of-the-art universal image restoration methods, including AirNet~\cite{airnet}, DGUNet~\cite{dgunet}, TransWeather~\cite{transweather}, PromptIR~\cite{promptir}, DiffIR~\cite{diffir}, DiffUIR~\cite{DiffUIR}, DA-CLIP~\cite{DA-CLIP}, AutoDIR~\cite{autodir}, FoundIR~\cite{foundir}, and DiT4SR~\cite{dit4sr}.
Except for methods marked with an asterisk (*) in Table~\ref{tab:foundir_dataset} and Table~\ref{tab: public_benchmark}, all compared approaches are retrained on our training set for a fair comparison. 
Specifically, for PromptIR and DiffUIR, we provide results from both officially released weights and our retrained versions to ensure comprehensive evaluation. For AutoDIR, only official weights are reported due to the unavailability of its training code.
{\flushleft \textbf{Evaluation on the FoundIR dataset.}} 
Table~\ref{tab:foundir_dataset} shows that the proposed method achieves competitive performance on reference-based metrics. 
For example, it obtains the best LPIPS scores on Noise, Haze, and Low-light degradations, demonstrating favorable fidelity preservation. 
Furthermore, our approach achieves the highest results on most no-reference metrics, such as NIQE, LIQE, MUSIQ, and CLIPIQA, significantly outperforming existing methods.
This indicates that our method produces visually more natural and perceptually better results.
Figure~\ref{fig: result_1}--\ref{fig: result_3} show visual comparisons on a variety of degradation tasks. 
For deblurring (Figure~\ref{fig: result_1}), our method effectively removes motion blur while preserving sharp edges and fine textures, resulting in clear and structurally faithful reconstructions. 
In dehazing (Figure~\ref{fig: result_2}), our approach successfully eliminates haze and recovers natural color and contrast, producing visually plausible and detailed results without color distortion or residual artifacts. 
For low-light enhancement (Figure~\ref{fig: result_3}), where input images suffer from severe illumination degradation, competing methods (Figure~\ref{fig: result_3}(c)--(g)) exhibit noticeable detail loss, over-smoothing, or unnatural brightness. 
In contrast, our method generates a significantly more detailed and visually pleasing result, with well-restored textures, accurate color reproduction, and minimal artifacts, demonstrating its enhanced capability in handling extreme low-light conditions.

These visual comparisons highlight the favorable generalization and restoration quality of our approach across diverse and challenging real-world scenarios.
More comparative results are provided in the supplementary material.

{\flushleft \textbf{Evaluation on additional public benchmarks.}}
We also evaluate our method on four additional public benchmarks.
Table~\ref{tab: public_benchmark} shows that our method achieves the best results on most no-reference metrics, indicating better perceptual quality, and remains within a close margin of the top performance on reference-based metrics, attesting to reliable fidelity preservation.
Visual comparisons on these datasets are provided in the supplementary material.

\subsection{Analysis and Discussion}
\label{sec:ablation}
To further demonstrate the effectiveness of the proposed method, we conduct ablation studies on the FoundIR dataset~\cite{foundir}.
%
%
We report the average of reference-based and no-reference metrics across all degradation tasks to provide an overall fidelity and perceptual quality assessment.

\begin{table}[!t]
    \caption{Ablation study on the effectiveness of the Intra-MoE module, comparing performance with and without Intra-MoE on the FoundIR dataset~\cite{foundir}. The variant with Intra-MoE achieves consistently better results, validating the importance of Intra-MoE.}

   \vspace{-3mm}
   \label{tab: ablation_moe}
\scriptsize

 \centering
\setlength{\tabcolsep}{3pt}
 \begin{tabular}{l|cccccc}
    \toprule
    Methods    &LPIPS~$\downarrow$&FID~$\downarrow$   &NIQE~$\downarrow$   &LIQE~$\uparrow$ & MUSIQ~$\uparrow$ &CLIPIQA~$\uparrow$\\
    \hline
  w/o Intra-MoE  & 0.1873    &34.5619  & 4.7352   &2.3143   &50.5890   &0.5511  \\
  w/ Intra-MoE    &\textbf{0.1636}&\textbf{26.8815}  &\textbf{4.2502}   &\textbf{2.9133}   &\textbf{53.4975}   &\textbf{0.5763}  \\
 \bottomrule
  \end{tabular}
\vspace{-6mm}
\end{table}

\begin{figure}[!t]\footnotesize
\centering
\begin{tabular}{cccc}
\hspace{-4mm}
\includegraphics[width=0.115\textwidth]{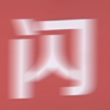} &\hspace{-4mm}
\includegraphics[width=0.115\textwidth]{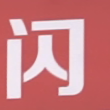} &\hspace{-4mm}
\includegraphics[width=0.115\textwidth]{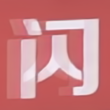} &\hspace{-4mm}
\includegraphics[width=0.115\textwidth]{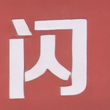} \\
\hspace{-3mm}(a) Input  &\hspace{-4mm}  (b) GT  &\hspace{-4mm} (c) w/o Intra-MoE  &\hspace{-4mm} (d) w/ Intra-MoE\\
\end{tabular}
\vspace{-2mm}
\caption{Effect of the Intra-MoE. The model with Intra-MoE produces a sharper and clearer result.}
\label{fig: MoE}
\vspace{-8mm}
\end{figure}

{\flushleft \textbf{Effect of the Intra-MoE.}} 
We first ablate the contribution of Intra-MoE.
To this end, we retain the same four parallel branches—spatial self-attention, channel self-attention, Swin attention, and SE attention—but replace the MoE module in each branch with a single fixed expert (i.e., removing intra-structure expert diversity and Top-k routing). 
The fusion across structures remains adaptive via the top-level gating network.
As shown in Table~\ref{tab: ablation_moe}, this variant without intra-structure MoE modules achieves lower MUSIQ and CLIPIQA than our full model, indicating worse perceptual quality. 
Visual results in Figure~\ref{fig: MoE} show that this baseline produces over-smoothed textures and fails to recover fine details under severe motion blur, especially in edge regions. This confirms that the MoE mechanism within each structure enables fine-grained adaptation to intra-degradation variations, leading to more realistic and detail-preserving restoration.

\begin{table}[!t]
    \caption{Ablation study on structural heterogeneity in Inter-MoE. Heterogeneous Inter-MoE (fusing all four attention types) outperforms single-structure variants across all metrics on FoundIR dataset~\cite{foundir}, proving that combining multiple attention mechanisms is essential for all-in-one restoration.
  }
   \vspace{-3mm}
   \label{tab: ablation_heterogeneity}
\scriptsize

 \centering
\setlength{\tabcolsep}{2pt}

 \begin{tabular}{l|ccccccc}
    \toprule
    Methods        &LPIPS~$\downarrow$   &FID~$\downarrow$        &NIQE~$\downarrow$     &LIQE~$\uparrow$    & MUSIQ~$\uparrow$      &CLIPIQA~$\uparrow$\\
    \hline
(a) Spatial-only   &0.1924  &35.8375    &4.3574   &2.6470        &52.4738     &0.5673  \\
(b) Channel-only   &0.1801  &31.3849    &4.2863   &2.5394        &53.4217     &0.5526  \\
(c) Swin-only      &0.2199  &38.1274    &4.9472   &2.5539        &51.6261     &0.5395\\
(d) SE-only        &0.2577  &40.3904    &5.1836   &2.1044        &49.1557     &0.4972\\
(e) Heterogeneity  &\textbf{0.1636} &\textbf{26.8815}  &\textbf{4.2502}   &\textbf{2.9133}   &\textbf{53.4975}   &\textbf{0.5763}    \\
 \bottomrule 
  \end{tabular}
\vspace{-6mm}
\end{table}
\vspace{-2mm}
{\flushleft \textbf{Effect of Inter-MoE.}} 
We further ablate the benefit of structural heterogeneity in Inter-MoE by evaluating variants that use homogeneous attention types across all expert groups. 
Specifically, we replace the diverse expert set, which includes spatial, channel, Swin, and SE attention, with identical modules. 
We conducted experiments on all four homogeneous configurations: Spatial-only, Channel-only, Swin-only, and SE-only expert groups, each while retaining the Intra-MoE design within every group.
As shown in Table~\ref{tab: ablation_heterogeneity}, our heterogeneous model consistently outperforms all variants, including spatial only, channel only, Swin only, and SE only configurations. 
This result demonstrates that a single type of attention mechanism is insufficient to handle diverse degradation patterns effectively.
%
%
By adaptively integrating multiple attention types, the proposed method successfully leverages their complementary inductive biases, enabling robust and perceptually convincing restoration across a wide variety of degradation scenarios. 
Visual comparisons of the ablation results are provided in the supplementary material.
%


\vspace{-3mm}
\begin{table}[H]
  \caption{Ablation studies of the Router Design, Dense Inter-MoE + Sparse Intra-MoE outperforms alternative configurations on FoundIR dataset~\cite{foundir}.
  }
   \vspace{-3mm}
   \label{tab: ablation_router}
\scriptsize

 \centering
\setlength{\tabcolsep}{0.5pt}
 \begin{tabular}{l|cccccc}
    \toprule
    Methods       &LPIPS~$\downarrow$ &FID~$\downarrow$ &NIQE~$\downarrow$   &LIQE~$\uparrow$& MUSIQ~$\uparrow$ &CLIPIQA~$\uparrow$\\
    \hline
Sparse Inter + Sparse Intra  &0.2457  &43.5841     &5.4490   &2.0107   &48.8349  &0.5183  \\
Sparse Inter + Dense Intra   &0.2348  &44.8341     &5.1905   &2.3049   &46.4920  &0.5532  \\
Dense Inter + Sparse Intra   &\textbf{0.1636} &\textbf{26.8815}  &\textbf{4.2502}   &\textbf{2.9133}   &\textbf{53.4975}   &\textbf{0.5763}      \\
 \bottomrule
  \end{tabular}
\vspace{-4mm}
\end{table}
{\flushleft \textbf{Router design in Inter-MoE and Intra-MoE.}} 
To validate the effectiveness of our routing strategy, we conduct ablation studies on the router design at both Inter-MoE and Intra-MoE levels. 
Specifically, we evaluate two alternative configurations: (1) replacing the dense fusion in Inter-MoE with sparse routing (selecting only one expert group) while keeping the original sparse routing in Intra-MoE unchanged, and (2) using sparse routing in Inter-MoE combined with dense fusion (utilizing all sub-experts equally) in Intra-MoE.
Adopting dense fusion at both levels would incur prohibitive computational overhead and render training infeasible, underscoring the necessity of our dense-to-sparse hybrid design.

Table~\ref{tab: ablation_router} shows that both alternative configurations lead to performance degradation across multiple metrics.
Sparse Inter-MoE routing curtails the integration of complementary structural priors, leading to degraded restoration performance.
Meanwhile, adopting dense fusion in Intra-MoE not only increases computational overhead but also fails to achieve significant performance improvement.
Our approach employs dense routing in Inter-MoE for comprehensive integration of diverse structural features and sparse routing in Intra-MoE for efficient processing within each expert group.
Visual comparisons of the ablation results are provided in the supplementary material.

{\flushleft \textbf{Analysis of adaptive routing weights in Inter-MoE.}} 
To validate that the dense router in our Inter-MoE learns meaningful and degradation-aware routing strategies, we analyze the average adaptive weights assigned to each expert group across various degradation types. 
As shown in Table~\ref{tab:routing_weights}, the routing distributions reveal systematic preferences: for degradations involving spatial distortions (blur and rain), the router prioritizes spatial and Swin attention to handle geometric corrections; for atmospheric effects (haze and low-light), it emphasizes channel and SE attention for illumination and transmission modeling; while for noise corruption, it balances spatial and channel attention to jointly address artifact removal and feature preservation. 
These consistent patterns demonstrate that our router effectively adapts to different degradation characteristics by dynamically activating complementary experts rather than employing a fixed fusion strategy.

\begin{table}[!t]
\centering
\caption{Average adaptive weights across degradation types. Values represent normalized mean routing weights.}
   \vspace{-3mm}

\label{tab:routing_weights}
\scriptsize
\setlength{\tabcolsep}{7.5pt}
\begin{tabular}{l|cccccc}
\hline
{Attention Type} & {Blur} & {Noise} & {Haze} & {Rain} & {Low-light} \\
\hline
Spatial    &0.3622    &0.3953    &0.1630    &0.3321    &0.2081   \\
Channel    &0.1399    &0.2784    &0.4008    &0.1406    &0.3716    \\
Swin       &0.3592    &0.1514    &0.2006    &0.3673    &0.1423    \\
SE         &0.1387    &0.1749    &0.2356    &0.1600    &0.2780    \\
\hline
\end{tabular}
\vspace{-7mm}
\end{table}
\label{sec:routing_analysis}

\section{Conclusion}
\label{sec:conclusion}

In this paper, we propose MiM-DiT, an effective all-in-one image restoration framework that integrates a hierarchical Mixture-of-Experts architecture with a pretrained Diffusion Transformer. 
By organizing heterogeneous attention modules into a dual-level MoE structure, our model dynamically adapts to diverse degradations while preserving the favorable generative priors of the diffusion backbone.
Within a unified trainable framework, we jointly optimize expert networks, gating modules, and the conditioning pathway to achieve end-to-end learning of specialized representations.
%
%
The interplay between structural diversity and adaptive routing allows the network to better disentangle degradation effects from intrinsic image content, yielding high-fidelity and perceptually realistic restoration.
Extensive experiments demonstrate that the proposed method performs favorably against the state-of-the-art
approaches on diverse image restoration tasks.

{
    \small
    \bibliographystyle{ieeenat_fullname}
    \bibliography{main}

@String(IJCV  = {IJCV})

@String(CVPR  = {CVPR})

@String(ICCV  = {ICCV})

@String(ECCV  = {ECCV})

@String(TIP   = {IEEE TIP})

@String(ICIP  = {ICIP})

@String(ICLR  = {ICLR})

@String(AAAI = {AAAI})

@inproceedings{dualCNNconfernce,
  author       = {Jinshan Pan and
                  Sifei Liu and
                  Deqing Sun and
                  Jiawei Zhang and
                  Yang Liu and
                  Jimmy S. J. Ren and
                  Zechao Li and
                  Jinhui Tang and
                  Huchuan Lu and
                  Yu{-}Wing Tai and
                  Ming{-}Hsuan Yang},
  title        = {Learning Dual Convolutional Neural Networks for Low-Level Vision},
  booktitle    = {CVPR},
  pages        = {3070--3079},
  year         = {2018}
}

@article{dualCNN,
  author       = {Jinshan Pan and
                  Deqing Sun and
                  Jiawei Zhang and
                  Jinhui Tang and
                  Jian Yang and
                  Yu{-}Wing Tai and
                  Ming{-}Hsuan Yang},
  title        = {Dual Convolutional Neural Networks for Low-Level Vision},
  journal      = {IJCV},
  volume       = {130},
  number       = {6},
  pages        = {1440--1458},
  year         = {2022}
}

@inproceedings{Adam,
  title={Adam: A Method for Stochastic Optimization},
  author={Kingma, Diederik P and Ba, Jimmy},
  booktitle={ICLR},
  year={2015}
}

@inproceedings{Restormer,
    title={Restormer: Efficient Transformer for High-Resolution Image Restoration}, 
    author={Syed Waqas Zamir and Aditya Arora and Salman Khan and Munawar Hayat 
            and Fahad Shahbaz Khan and Ming-Hsuan Yang},
    booktitle={CVPR},
    year={2022}
}

@inproceedings{NAFNet,
  title={Simple Baselines for Image Restoration},
  author={Chen, Liangyu and Chu, Xiaojie and Zhang, Xiangyu and Sun, Jian},
  booktitle={ECCV},
  year={2022}
}

@inproceedings{MPRNet,
    title={Multi-Stage Progressive Image Restoration},
    author={Syed Waqas Zamir and Aditya Arora and Salman Khan and Munawar Hayat
            and Fahad Shahbaz Khan and Ming-Hsuan Yang and Ling Shao},
    booktitle={CVPR},
    year={2021}
}

@inproceedings{Transformer,
  title={Attention is All you Need},
  author={Ashish Vaswani and Noam Shazeer and Niki Parmar and Jakob Uszkoreit and Llion Jones and Aidan N. Gomez and Lukasz Kaiser and Illia Polosukhin},
  booktitle={NeurIPS},
  year={2017}
}

@inproceedings{Swin,
  title={Swin Transformer: Hierarchical Vision Transformer using Shifted Windows},
  author={Liu, Ze and Lin, Yutong and Cao, Yue and Hu, Han and Wei, Yixuan and Zhang, Zheng and Lin, Stephen and Guo, Baining},
  booktitle={ICCV},
  year={2021}
}

@article{physicgan,
    author  = {Jinshan Pan and Jiangxin Dong and Yang Liu and Jiawei Zhang and Jimmy S. J. Ren and Jinhui Tang and Yu-Wing Tai  and Ming-Hsuan Yang},
    title   = {Physics-based generative adversarial models for image restoration and beyond},
    journal = {IEEE TPAMI},  
    year    = {2021},
    volume  = {43},
    number  = {7},
    pages   = {2449-2462}
}

@inproceedings{senet,
  title={Squeeze-and-excitation networks},
  author={Hu, Jie and Shen, Li and Sun, Gang},
  booktitle={CVPR},
  year={2018}
}

@inproceedings{faithdiff,
  title={Faithdiff: Unleashing diffusion priors for faithful image super-resolution},
  author={Chen, Junyang and Pan, Jinshan and Dong, Jiangxin},
  booktitle={CVPR},
  year={2025}
}

@inproceedings{promptir,
  title={PromptIR: Prompting for All-in-One Image Restoration},
  author={Potlapalli, Vaishnav and Zamir, Syed Waqas and Khan, Salman and Khan, Fahad},
  booktitle={NeurIPS},
  year={2023}
}

@inproceedings{moceir,
  title={Complexity experts are task-discriminative learners for any image restoration},
  author={Zamfir, Eduard and Wu, Zongwei and Mehta, Nancy and Tan, Yuedong and Paudel, Danda Pani and Zhang, Yulun and Timofte, Radu},
  booktitle={CVPR},
  year={2025}
}

@inproceedings{autodir,
  title={Autodir: Automatic all-in-one image restoration with latent diffusion},
  author={Jiang, Yitong and Zhang, Zhaoyang and Xue, Tianfan and Gu, Jinwei},
  booktitle={ECCV},
  year={2024}
}

@inproceedings{instructir,
  title={Instructir: High-quality image restoration following human instructions},
  author={Conde, Marcos V and Geigle, Gregor and Timofte, Radu},
  booktitle={ECCV},
  year={2024},
}

@inproceedings{foundir,
      title={FoundIR: Unleashing Million-scale Training Data to Advance Foundation Models for Image Restoration},
      author={Li, Hao and Chen, Xiang and Dong, Jiangxin and Tang, Jinhui and Pan, Jinshan},
      booktitle={ICCV},
      year={2025}
}

@inproceedings{DA-CLIP,
  title={Controlling vision-language models for universal image restoration},
  author={Luo, Ziwei and Gustafsson, Fredrik K and Zhao, Zheng and Sj{\"o}lund, Jens and Sch{\"o}n, Thomas B},
  booktitle={ICLR},
  year={2023}
}

@inproceedings{DiffUIR,
  title={Selective hourglass mapping for universal image restoration based on diffusion model},
  author={Zheng, Dian and Wu, Xiao-Ming and Yang, Shuzhou and Zhang, Jian and Hu, Jian-Fang and Zheng, Wei-Shi},
  booktitle={CVPR},
  year={2024}
}

@inproceedings{diffir,
  title={Diffir: Efficient diffusion model for image restoration},
  author={Xia, Bin and Zhang, Yulun and Wang, Shiyin and Wang, Yitong and Wu, Xinglong and Tian, Yapeng and Yang, Wenming and Van Gool, Luc},
  booktitle={CVPR},
  year={2023}
}

@inproceedings{airnet,
  title={All-in-one image restoration for unknown corruption},
  author={Li, Boyun and Liu, Xiao and Hu, Peng and Wu, Zhongqin and Lv, Jiancheng and Peng, Xi},
  booktitle={CVPR},
  year={2022}
}

@inproceedings{aio1,
  title={All in one bad weather removal using architectural search},
  author={Li, Ruoteng and Tan, Robby T and Cheong, Loong-Fah},
  booktitle={CVPR},
  year={2020}
}

@inproceedings{cat,
  title={Cross aggregation transformer for image restoration},
  author={Chen, Zheng and Zhang, Yulun and Gu, Jinjin and Kong, Linghe and Yuan, Xin and others},
  booktitle={NeurIPS},
  year={2022}
}

@article{Prores,
  title={Prores: Exploring degradation-aware visual prompt for universal image restoration},
  author={Ma, Jiaqi and Cheng, Tianheng and Wang, Guoli and Zhang, Qian and Wang, Xinggang and Zhang, Lefei},
  journal={arXiv preprint arXiv:2306.13653},
  year={2023}
}

@inproceedings{DDPM,
  title={Denoising diffusion probabilistic models},
  author={Ho, Jonathan and Jain, Ajay and Abbeel, Pieter},
  booktitle={NeurIPS},
  year={2020}
}

@inproceedings{cdm,
  title={Multiscale structure guided diffusion for image deblurring},
  author={Ren, Mengwei and Delbracio, Mauricio and Talebi, Hossein and Gerig, Guido and Milanfar, Peyman},
  booktitle={CVPR},
  year={2023}
}

@inproceedings{resshift,
  title={Resshift: Efficient diffusion model for image super-resolution by residual shifting},
  author={Yue, Zongsheng and Wang, Jianyi and Loy, Chen Change},
  booktitle={NeurIPS},
  year={2024}
}

@inproceedings{diffbir,
  title={Diffbir: Toward blind image restoration with generative diffusion prior},
  author={Lin, Xinqi and He, Jingwen and Chen, Ziyan and Lyu, Zhaoyang and Dai, Bo and Yu, Fanghua and Qiao, Yu and Ouyang, Wanli and Dong, Chao},
  booktitle={ECCV},
  year={2024},
}

@inproceedings{pasd,
    title={Pixel-Aware Stable Diffusion for Realistic Image Super-Resolution and Personalized Stylization},
    author={Yang, Tao and Wu, Rongyuan and Ren, Peiran and Xie, Xuansong and Zhang, Lei},
    booktitle={ECCV},
    year={2024}
}

@inproceedings{sparse,
  title={Blind deconvolution using a normalized sparsity measure},
  author={Krishnan, Dilip and Tay, Terence and Fergus, Rob},
  booktitle={CVPR},
  year={2011},
}

@inproceedings{sd,
  title={High-resolution image synthesis with latent diffusion models},
  author={Rombach, Robin and Blattmann, Andreas and Lorenz, Dominik and Esser, Patrick and Ommer, Bj{\"o}rn},
  booktitle={CVPR},
  year={2022}
}

@inproceedings{4krd,
  title={Multi-scale separable network for ultra-high-definition video deblurring},
  author={Deng, Senyou and Ren, Wenqi and Yan, Yanyang and Wang, Tao and Song, Fenglong and Cao, Xiaochun},
  booktitle={ICCV},
  year={2021}
}

@article{realrain,
  title={Toward real-world single image deraining: A new benchmark and beyond},
  author={Li, Wei and Zhang, Qiming and Zhang, Jing and Huang, Zhen and Tian, Xinmei and Tao, Dacheng},
  journal={arXiv preprint arXiv:2206.05514},
  year={2022}
}

@inproceedings{hazerd,
  title={Hazerd: an outdoor scene dataset and benchmark for single image dehazing},
  author={Zhang, Yanfu and Ding, Li and Sharma, Gaurav},
  booktitle={ICIP},
  year={2017},
}

@InProceedings{uhdll,
    author = {Li, Chongyi and Guo, Chun-Le and Zhou, Man and Liang, Zhexin and Zhou, Shangchen and Feng, Ruicheng and Loy, Chen Change},
    title = {Embedding Fourier for Ultra-High-Definition Low-Light Image Enhancement},
    booktitle = {ICLR},
    year = {2023}
}

@article{NIQE,
  author={Zhang, Lin and Zhang, Lei and Bovik, Alan C.},
  journal={{IEEE} TIP}, 
  title={A Feature-Enriched Completely Blind Image Quality Evaluator}, 
  year={2015},
  volume={24},
  number={8},
  pages={2579-2591},
}

@INPROCEEDINGS{MUSIQ,
  author={Ke, Junjie and Wang, Qifei and Wang, Yilin and Milanfar, Peyman and Yang, Feng},
  booktitle={ICCV}, 
  title={MUSIQ: Multi-scale Image Quality Transformer}, 
  year={2021},
}

@inproceedings{CLIP-IQA,
    author = {Wang, Jianyi and Chan, Kelvin CK and Loy, Chen Change},
    title = {Exploring CLIP for Assessing the Look and Feel of Images},
    booktitle = {AAAI},
    year = {2023}
}

@inproceedings{FID,
  title={Gans trained by a two time-scale update rule converge to a local nash equilibrium},
  author={Martin, Heusel and Hubert, Ramsauer and Thomas, Unterthiner and Bernhard, Nessler and Sepp, Hochreiter},
  booktitle={NerrIPS},
  year={2017},
}

@inproceedings{lpips,
  title={The unreasonable effectiveness of deep features as a perceptual metric},
  author={Zhang, Richard and Isola, Phillip and Efros, Alexei A and Shechtman, Eli and Wang, Oliver},
  booktitle={CVPR},
  year={2018}
}

@inproceedings{dgunet,
  title={Deep generalized unfolding networks for image restoration},
  author={Mou, Chong and Wang, Qian and Zhang, Jian},
  booktitle={CVPR},
  year={2022}
}

@inproceedings{transweather,
  title={Transweather: Transformer-based restoration of images degraded by adverse weather conditions},
  author={Valanarasu, Jeya Maria Jose and Yasarla, Rajeev and Patel, Vishal M},
  booktitle={CVPR},
  year={2022}
}

@inproceedings{dit4sr,
  title={DiT4SR: Taming Diffusion Transformer for Real-World Image Super-Resolution},
  author={Duan, Zheng-Peng and Zhang, Jiawei and Jin, Xin and Zhang, Ziheng and Xiong, Zheng and Zou, Dongqing and Ren, Jimmy and Guo, Chun-Le and Li, Chongyi},
  booktitle={ICCV},
  year={2025}
}

@inproceedings{dit,
  title={Scalable diffusion models with transformers},
  author={Peebles, William and Xie, Saining},
  booktitle={ICCV},
  year={2023}
}

@inproceedings{RectifiedFlow,
  title={Flow straight and fast: Learning to generate and transfer data with rectified flow},
  author={Liu, Xingchao and Gong, Chengyue and Liu, Qiang},
  booktitle={ICLR},
  year={2022}
}

@inproceedings{flowmatching,
  title={Flow matching for generative modeling},
  author={Lipman, Yaron and Chen, Ricky TQ and Ben-Hamu, Heli and Nickel, Maximilian and Le, Matt},
  booktitle={ICLR},
  year={2022}
}

@article{unirestorer,
  title={UniRestorer: Universal Image Restoration via Adaptively Estimating Image Degradation at Proper Granularity},
  author={Lin, Jingbo and Zhang, Zhilu and Li, Wenbo and Pei, Renjing and Xu, Hang and Zhang, Hongzhi and Zuo, Wangmeng},
  journal={arXiv preprint arXiv:2412.20157},
  year={2024}
}

@article{moe,
  title={Adaptive mixtures of local experts},
  author={Jacobs, Robert A and Jordan, Michael I and Nowlan, Steven J and Hinton, Geoffrey E},
  journal={Neural computation},
  volume={3},
  number={1},
  pages={79--87},
  year={1991},
  publisher={MIT Press}
}

@inproceedings{hidiff,
  title={Hierarchical Integration Diffusion Model for Realistic Image Deblurring}, 
  author={Chen, Zheng and Zhang, Yulun and Ding, Liu and Bin, Xia and Gu, Jinjin and Kong, Linghe and Yuan, Xin},
  booktitle={NeurIPS},
  year={2023}
}

@article{aio_lvm1,
  title={Controlling vision-language models for universal image restoration},
  author={Luo, Ziwei and Gustafsson, Fredrik K and Zhao, Zheng and Sj{\"o}lund, Jens and Sch{\"o}n, Thomas B},
  journal={arXiv preprint arXiv:2310.01018},
  year={2023}
}

@inproceedings{uniprocessor,
  title={Uniprocessor: a text-induced unified low-level image processor},
  author={Duan, Huiyu and Min, Xiongkuo and Wu, Sijing and Shen, Wei and Zhai, Guangtao},
  booktitle={ECCV},
  year={2024},
}

@inproceedings{liqe,
  title={Blind image quality assessment via vision-language correspondence: A multitask learning perspective},
  author={Zhang, Weixia and Zhai, Guangtao and Wei, Ying and Yang, Xiaokang and Ma, Kede},
  booktitle={CVPR},
  year={2023}
}
}


\end{document}